\documentclass[letterpaper, 10 pt, conference]{ieeeconf}  

\IEEEoverridecommandlockouts                              

\usepackage{graphics} 
\usepackage{epsfig} 
\usepackage{graphicx}
\usepackage{amsmath} 
\usepackage{amssymb}  
\usepackage{cite}
\usepackage{colortbl}
\usepackage{threeparttable}
\usepackage{booktabs}
\usepackage{color}
\usepackage{array}
\usepackage{multirow}

\title{\LARGE \bf
	RangeRCNN: Towards Fast and Accurate 3D Object Detection \\with Range Image Representation 
}

\author{Zhidong Liang$^{1}$, Ming Zhang$^{1}$, Zehan Zhang$^{1}$, Xian Zhao$^{1}$, and Shiliang Pu$^{1}$
	\thanks{$^{1}$Zhidong Liang, Ming Zhang, Zehan Zhang, Xian Zhao, Shiliang Pu are with Hikvision Research Institute, Hangzhou Hikvision Digital Technology Co. Ltd, China~(e-mail: liangzhidong@hikvision.com; zhangming15@hikvision.com; zhangzehan@hikvision.com; zhaoxian@hikvision.com; pushiliang.hri@hikvision.com)}
}

\begin{document}

	\maketitle
	\thispagestyle{empty}
	\pagestyle{empty}

	\begin{abstract}
		We present RangeRCNN, a novel and effective 3D object detection framework based on the range image representation. Most existing methods are voxel-based or point-based. Though several optimizations have been introduced to ease the sparsity issue and speed up the running time, the two representations are still computationally inefficient. Compared to them, the range image representation is dense and compact which can exploit powerful 2D convolution. Even so, the range image is not preferred in 3D object detection due to scale variation and occlusion. In this paper, we utilize the dilated residual block~(DRB) to better adapt different object scales and obtain a more flexible receptive field. Considering scale variation and occlusion, we propose the RV-PV-BEV~(range view-point view-bird's eye view) module to transfer features from RV to BEV. The anchor is defined in BEV which avoids scale variation and occlusion. Neither RV nor BEV can provide enough information for height estimation; therefore, we propose a two-stage RCNN for better 3D detection performance. The aforementioned point view not only serves as a bridge from RV to BEV but also provides pointwise features for RCNN. Experiments show that RangeRCNN achieves state-of-the-art performance on the KITTI dataset and the Waymo Open dataset, and provides more possibilities for real-time 3D object detection. We further introduce and discuss the data augmentation strategy for the range image based method, which will be very valuable for future research on range image.
	\end{abstract}

	\section{INTRODUCTION}
	In recent years, 3D object detection has attracted increasing attention in many fields. The well-studied 2D object detection can only determine the object position in the 2D pixel space instead of the 3D physical space. However, the 3D information is extremely important for several applications, such as autonomous driving. Compared to 2D object detection, 3D object detection remains challenging since the point cloud is irregular and sparse. The suitable representation for the 3D point cloud is worthy of research.
	
	Existing methods are mostly divided into two categories: the grid-based representation and the point-based representation. The grid-based representation can be further classified into two classes: 3D voxels and 2D BEV~(bird's eye view). Such representations can utilize the 3D/2D convolution to extract features, but simultaneously suffer from the information loss of quantization. The 3D convolution is not efficient and practical in large outdoor scenes even though several optimizations have been proposed~\cite{yan2018second, graham20183d}. The 2D BEV suffers from more severe information loss than the 3D voxel which limits its performance. The point-based representation retains more information than the voxel-based methods. However, the point-based methods are generally inefficient when the number of points is large. Downsampling points can reduce the computation cost but simultaneously degrades the localization accuracy. In summary, neither of the two representations can retain all original information for feature extraction while being computationally efficient.
	
\begin{figure}[t]
		\centering
		\includegraphics[width=1.0\linewidth]{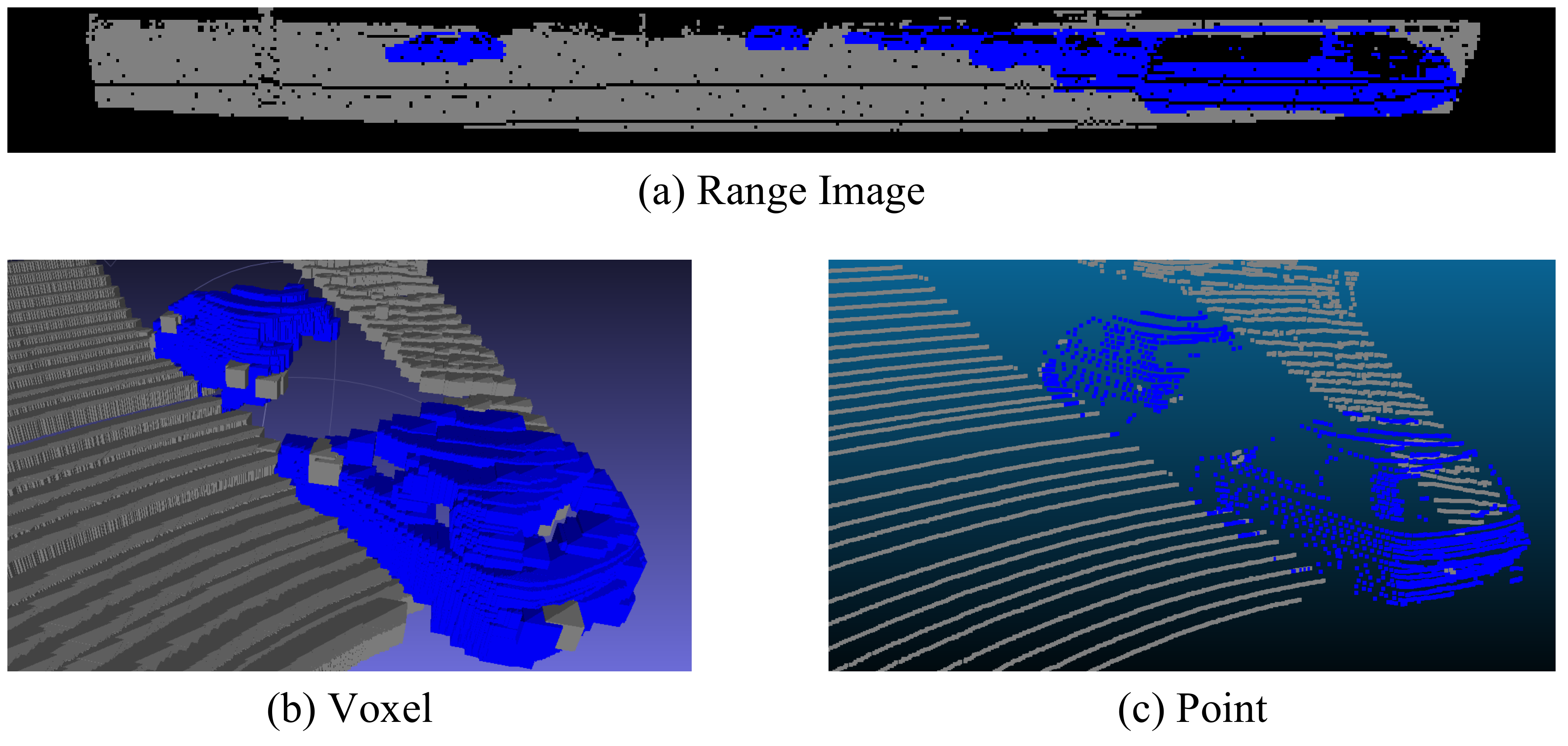}
		\caption{Different representations of point clouds. (a) Range image representation~(dense). Use 2D convolution to extract features. (b) 3D Voxel representation~(sparse). Use 3D convolution to extract features. (c) Point representation~(sparse). Use point-based convolution to extract features.}
		\label{intro}
	\end{figure}

	Although we mostly regard the point cloud as the raw data format, the range image is the native representation of the rotating LIDAR sensor~(e.g. Velodyne 64E, etc). It retains all original information without any loss. Beyond this, the dense and compact properties make it efficient to process. Fig.~\ref{intro} shows the three representations of point clouds. As a result, we consider it beneficial to extract features from the range image. Several methods~\cite{li2016vehicle, Meyer_2019_CVPR} directly operate on the range image but have not achieved performance similar to the voxel-based and point-based methods. \cite{Meyer_2019_CVPR} attributes the unsatisfactory performance to the small size of the KITTI dataset which makes it difficult to learn from the range image, and conducts the experimentation on their large private dataset to prove the effectiveness of the range image. In this paper, we present a range image based method and prove that the range image representation can also achieve state-of-the-art performance on the KITTI dataset. We further evaluate our method on the large-scale Waymo Open dataset, which provides the raw range images collected from their LIDAR sensors and achieve the top performance.
	
	\begin{figure*}[!tb]
		\centering
		\includegraphics[width=0.99\linewidth]{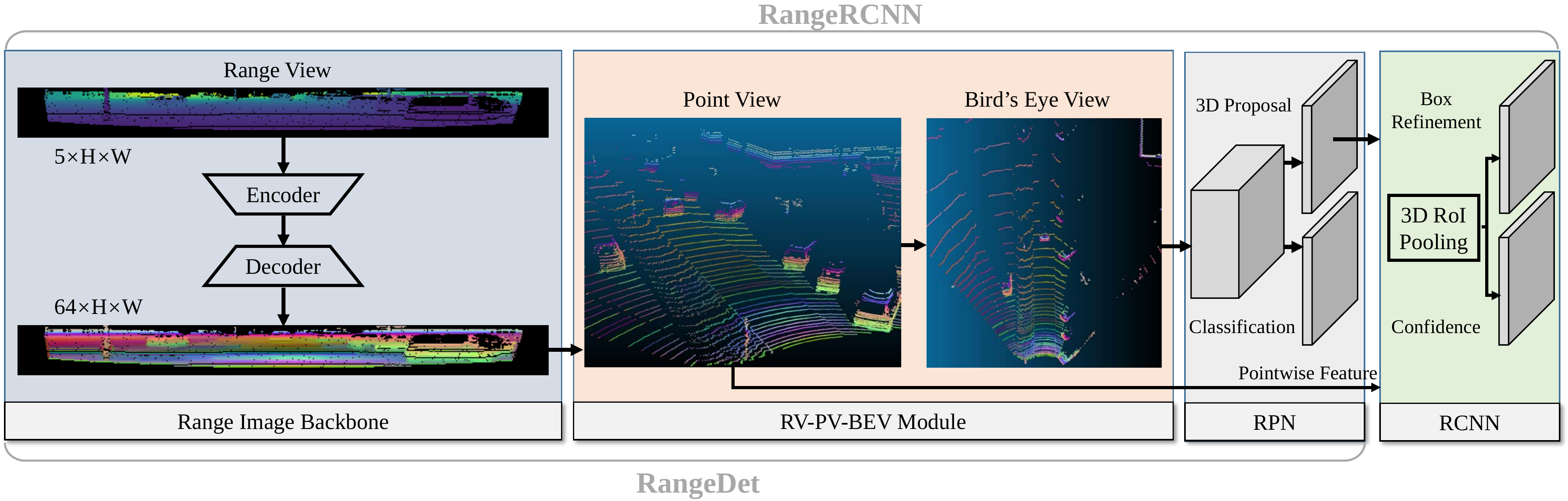}
		\caption{Illustration of the framework of RangeRCNN. The input range image is visualized using the pseudocolor according to the range channel. After the range image backbone, the high-level feature with 64 dimensions is extracted. We visualize it by t-SNE dimension reduction. The features extracted from the range image are transferred to the point view and the bird's eye view in turn. The region proposal network~(RPN) is used to generate 3D proposals from BEV. The 3D proposal and the pointwise feature are input into the 3D RoI pooling for proposal refinement. We name the one-stage network without RCNN RangeDet, and name the whole two-stage framework RangeRCNN.}
		\label{framework}
	\end{figure*}
	
	Though several advantages of the range image are pointed out above, its essential drawbacks are also obvious. The large scale variation makes it difficult to decide the anchor size in the range view, and the occlusion causes the bounding boxes to easily overlap with each other. These two issues do not exist in the 2D BEV space. Considering these properties, we propose a novel framework named RangeRCNN. First, we extract features from the range image for its compact and lossless representation. To better adapt the scale variation of the range image, we utilize the dilated residual block which uses the dilated convolution~\cite{chen2017deeplab} to achieve a more flexible receptive field. Then, we propose the RV-PV-BEV module to transfer the features extracted from the range view to the bird's eye view. Because the high-level features are effectively extracted, the influence of the quantization error caused by BEV is attenuated. The BEV mainly plays the role of anchor generation. Neither the range image nor the bird's eye image can explicitly supervise the height of the 3D bounding box. As a result, we propose to refine the 3D bounding box using a two-stage RCNN. The point view in the RV-PV-BEV module does not only serve as the bridge from RV to BEV, but also provides pointwise features for RCNN refinement.

	In summary, the key contributions of this paper are as follows:
	
	\begin{itemize}
		
		\item We propose the RangeRCNN framework which takes the range image as the initial input to extract dense and lossless features for fast and accurate 3D object detection.
		
		\item We design a 2D CNN utilizing dilated convolution to better adapt the flexible receptive field of the range image.
		
		\item We propose the RV-PV-BEV module for transferring the feature from the range view to the bird's eye view for easier anchor generation.
		
		\item We propose an end-to-end two-stage pipeline that utilizes a region convolutional neural network~(RCNN) for better height estimation. The whole network does not use 3D convolution or point-based convolution which makes it simple and efficient.
		
		\item Our proposed RangeRCNN achieves state-of-the-art performance on the competitive KITTI 3D detection benchmark and large-scale Waymo Open dataset.
		
	\end{itemize}

	\section{RELATED WORK}
	\subsection{3D Object Detection}
	\textbf{3D Object Detection with Grid-based Methods. }
	Most state-of-the-art methods in 3D object detection project the point clouds to the regular grids. \cite{chen2017mv3d, ku2018avod, Liang_2019_CVPR, liang2018deep} directly projects the original point clouds to the 2D bird's eye view to utilize the efficient 2D convolution for feature extraction. They also combine RGB images and other views for deep feature fusion. \cite{zhou2018voxel} is a pioneering work in 3D voxel-based object detection. Based on \cite{zhou2018voxel}, \cite{yan2018second} increases the efficiency of the 3D convolution using sparse optimization. Several following methods~\cite{shi2019part, he2020structure, shi2020pv} utilize the sparse operation~\cite{yan2018second,graham20183d} to develop more accurate detectors. For the real-time 3D object detector, \cite{Lang_2019_CVPR} proposes the pillar-based voxel to significantly improve the efficiency. However, the grid-based methods suffer from information loss in the stage of initial feature extraction. The sparsity issue of the point clouds also limits the effective receptive field of 3D convolution. These problems will become more serious when processing the large outdoor scene.

	\textbf{3D Object Detection with Point-based Methods. }Compared to the grid-based methods, the point-based methods are limited by the high computation cost in early research works. \cite{qi2018fpoint, wang2019fconv} project 2D bounding boxes to the 3D space to obtain 3D frustums and conduct the 3D object detection in each frustum. \cite{shi2019pointrcnn, yang2019std} directly process the whole point cloud using \cite{qi2017pointnet++} and generate proposals in a bottom-up manner. \cite{qi2019vote} introduces a vote-based 3D detector that is more suitable to process indoor scenes. \cite{shi2020point} uses a graph neural network for point cloud detection. \cite{yang20203dssd} proposes a fusion sampling strategy to speed up the point-based method.

	\textbf{3D Object Detection with Range Image. }Compared to the grid-based and point-based methods, fewer researchers utilize the range image in 3D object detection. \cite{chen2017mv3d} takes the range image as one of its inputs. \cite{li2016vehicle,Meyer_2019_CVPR} directly process the range image for 3D object detection. However, the range image based methods have not matched the performance of the grid-based or point-based methods on the popular KITTI dataset until now. \cite{Meyer_2019_CVPR} mentions that better results are achieved on the larger self-collected dataset. Recently, \cite{bewley2020range} has proposed the range conditioned dilated block for learning scale invariant features from the range image and achieves the top performance on the large-scale Waymo Open dataset. We think that the scale of the dataset is a secondary reason for low accuracy. The main reason is that the range image is a good choice for extracting initial features, but not a good choice for generating anchors. In this paper, we design a better framework utilizing the range image representation for 3D object detection.

	\subsection{Feature Learning on point clouds}
	Recently, feature learning on point clouds has been studied for different tasks, such as classification, detection and segmentation. Among these tasks, the 3D semantic segmentation task is chosen by many methods~\cite{qi2017pointnet,qi2017pointnet++,li2018pointcnn,graham20183d,wu2018squeezeseg} as the touchstone for evaluating the ability to extract features from point clouds. Early research was mainly focused on indoor scenes due to the lack of outdoor datasets. SemanticKITTI~\cite{behley2019semantickitti} is a recent semantic segmentation benchmark for autonomous driving scenes. In the benchmark, \cite{milioto2019rangenet++} proves the effectiveness of learning features from the range image for 3D semantic segmentation and simultaneously runs at a high speed. However, the value of the range image in the field of 3D object detection is far from being fully explored. We believe that the range image can also provide rich and useful information for the 3D object detection task.

	\section{METHOD DESCRIPTION}
	In section~\ref{sec_arch}, we present the network architecture of our method. In section~\ref{sec_backbone}, we introduce the backbone for extracting features from the range image. In section~\ref{sec_rpe}, we introduce the RV-PV-BEV module for transferring features between different views. In section~\ref{sec_pooling}, we utilize 3D RoI pooling to refine the generated proposals. In section~\ref{sec_loss}, we describe the loss function used in our network.

	\subsection{Network Architecture} \label{sec_arch}
	Our proposed network is illustrated in Fig.~\ref{framework}. The 3D point cloud is represented as the native range image, which is fed into an encoder-decoder 2D backbone to extract features efficiently and effectively. We upsample the deep feature map to the original resolution of the range image using a decoder in order to retain more spatial information. We then transfer features from the range image to each point. We do not extract features based on the point view using the point-based convolution~\cite{qi2017pointnet, qi2017pointnet++}. Actually, the point view has two functions. First, it serves as the bridge from the range image to the bird's eye image. Second, it provides the pointwise feature to the 3D RoI pooling module for refining the proposals generated by the region proposal network~(RPN). After obtaining the pointwise feature, we can easily obtain the BEV feature by projecting the 3D point to the x-y plane. Since we have effectively extracted high-level features from the range image, BEV mainly plays the role of proposal generation. This projection is therefore different from projecting the 3D point to BEV at the beginning, which is extremely dependent on the feature extraction from BEV. We use a simple RPN to generate proposals from BEV and refine the proposals using the 3D RoI pooling module. We name the one-stage network without RCNN pooling RangeDet, and name the two-stage framework RangeRCNN.

	\begin{figure}[!t]
		\centering
		\includegraphics[width=0.96\linewidth]{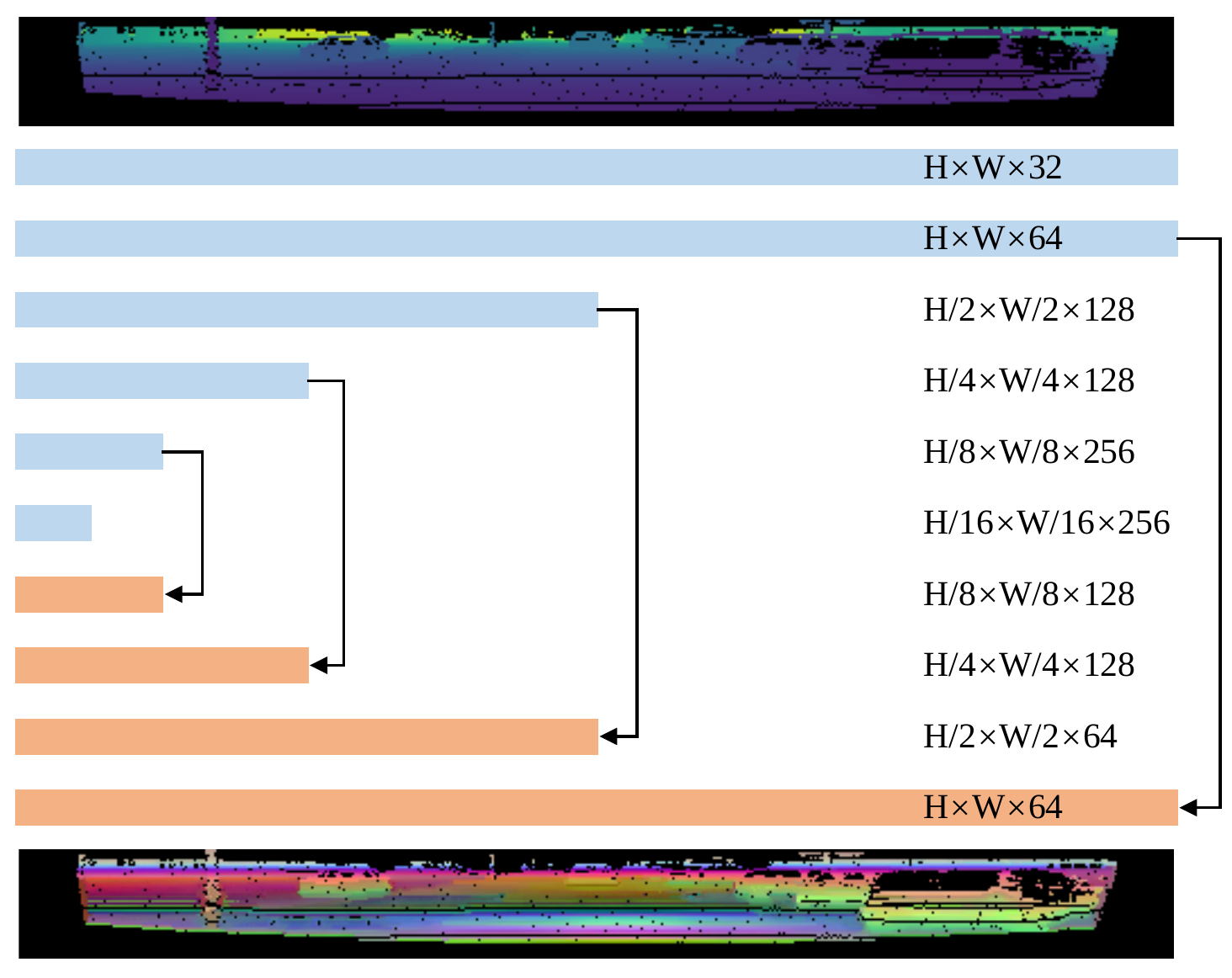}
		\caption{Illustration of the backbone for the range image. The backbone is an encoder-decoder structure for extracting high-level features.}
		\label{backbone}
	\end{figure}

	\subsection{Range Image Backbone} \label{sec_backbone}
	Most existing datasets such as KITTI provide the point cloud as the LIDAR data format, so we need to convert the points to the range image. As described in~\cite{milioto2019rangenet++}, the formula is as follows:
	
	\begin{equation}
		\begin{pmatrix}
		~u~ \\ ~v~
		\end{pmatrix}
		=
		\begin{pmatrix}
		~\frac{1}{2} [1 - arctan(y, x)\pi^{-1}] \times w   \\ [1 - (arcsin(z, r) + f_{down})f^{-1}] \times h
		\end{pmatrix}
		\label{conversion}
	\end{equation}
	
	where $(x,y,z)$ represents the point coordinate in the 3D space. $(u,v)$ is the pixel coordinate in the range image. $r=\sqrt{x^{2}+y^{2}+z^{2}}$ is the range of each point. $w$ and $h$ are the predefined width and height of the range image. $f = f_{up} + f_{down}$ is the vertical field-of-view of the LIDAR sensor. For each pixel position, we encode its range, coordinate, and intensity as the input channel. As a result, the size of the input range image is $5\times h\times w$. In~\cite{milioto2019rangenet++}, the categories of semantic segmentation are labeled for all points, so the range image contains the 360-degree information. The LIDAR used by the KITTI dataset is the Velodyne 64E LIDAR with 64 vertical channels. Each channel generates approximately 2000 points. In their task, therefore, $h$ and $w$ are respectively set as 64 and 2048. In the KITTI 3D detection task, the LIDAR and the camera are jointly calibrated. Only the objects in the front view of the camera are labeled; this view represents approximately 90 degrees of the whole scene. Additionally, some vertical channels are filtered by the FOV of the camera. We thus set $h=48$ and $w=512$, which is sufficient to contain the front view scene. The size of the range images used for the KITTI dataset is $5\times 48 \times 512$. The Waymo Open dataset provides range images with the size of $64 \times 2650$ as the LIDAR data format. Therefore, we can directly use it without any extra conversion.

	\begin{figure}[!t]
		\centering
		\includegraphics[width=1.0\linewidth]{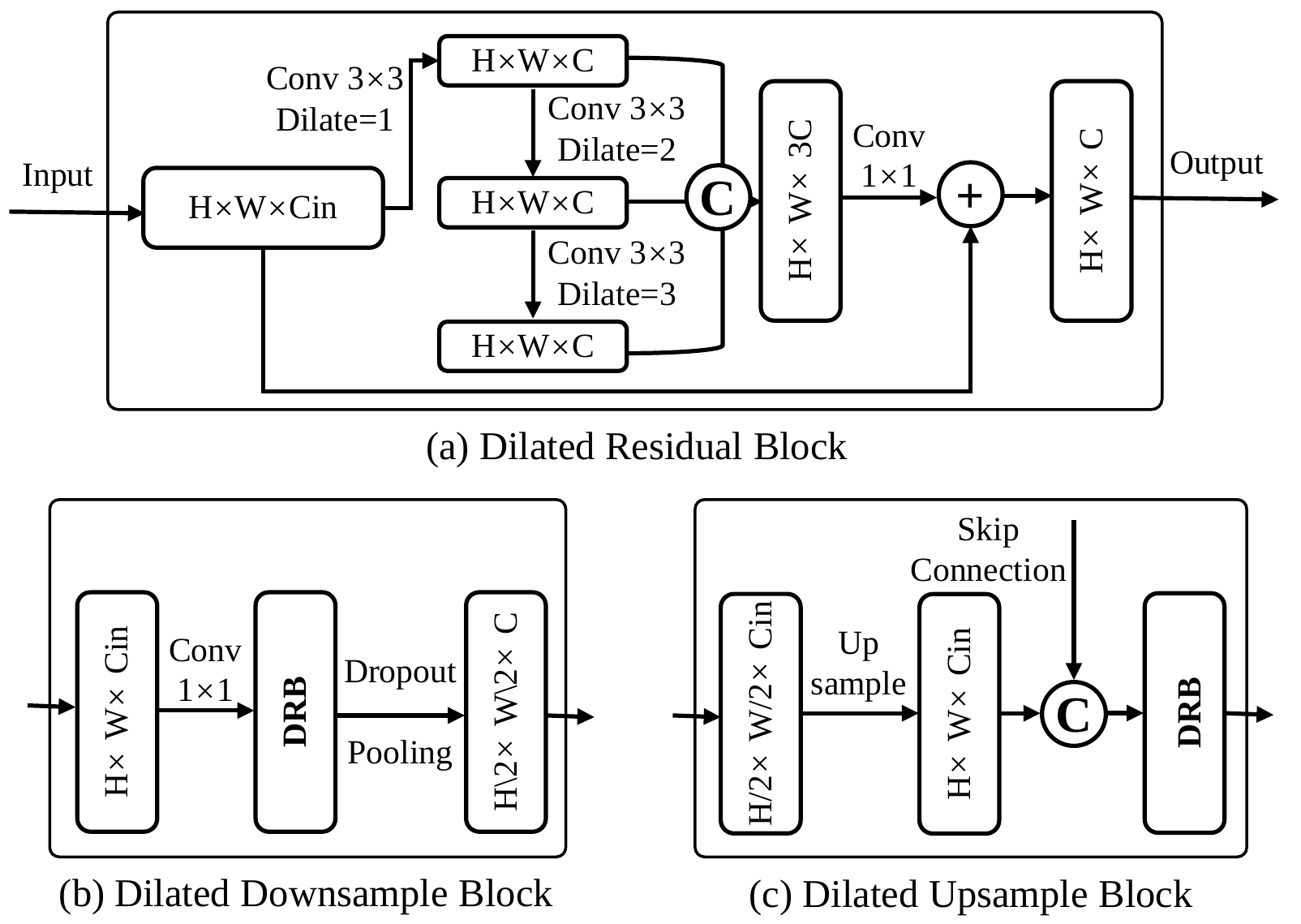}
		\caption{Illustration of the layer structure. (a) Dilated residual block~(DRB). (b) Dilated downsample block for the encoder. (c) Dilated upsample block for the decoder. }
		\label{DRB}
	\end{figure}

	The backbone for the range image is illustrated in Fig.~\ref{backbone}. The 2D backbone for extracting features from the range image is an encoder-decoder structure. The size of the feature map is given in the figure. The input is downsampled in the encoder by six dilated downsample blocks~(shown in Fig.~\ref{DRB}(b)), then gradually upsampled in the decoder by four corresponding dilated upsample blocks~(shown in Fig.~\ref{DRB}(c)). The range image provides dense and compact representation for utilizing the 2D CNN but simultaneously presents the scale variation issue. The scales of objects with different distances exhibit significant differences. To better adapt different scales and obtain a more flexible receptive field, we design the dilated residual block~(DRB), which inserts the dilated convolution into the normal residual block. In this block~(Fig.~\ref{DRB}(a)), three $3\times 3$ convolutions with different dilated rates~\{1, 2, 3\} are applied to extract features with different receptive fields. The outputs of the three dilated convolutions are concatenated followed by a $1\times 1$ convolution to fuse the features with different receptive fields. A residual connection is used to add the fused feature and the input feature. In the dilated downsample block~(Fig.~\ref{DRB}(b)), we first use a $1\times 1$ convolution to extract features across channels. Then the feature is input to DRB to extract features with a flexible receptive field. The dropout operation and the pooling operation are used for better generalization performance and downsampling the feature map, respectively. In the first two blocks of the encoder, we do not use the pooling operation. In the dilated upsample block~(Fig.~\ref{DRB}(c)), we first use the bilinear interpolation to upsample the feature map. A skip connection is introduced to concatenate the feature from the encoder. The concantenated feature is then input to DRB. The size of the final output features~(visualized in Fig.~\ref{backbone} by t-SNE) extracted from the range image is the same size as the input with 64 dimensions.

	\begin{figure}[!t]
		\centering
		\includegraphics[width=1.0\linewidth]{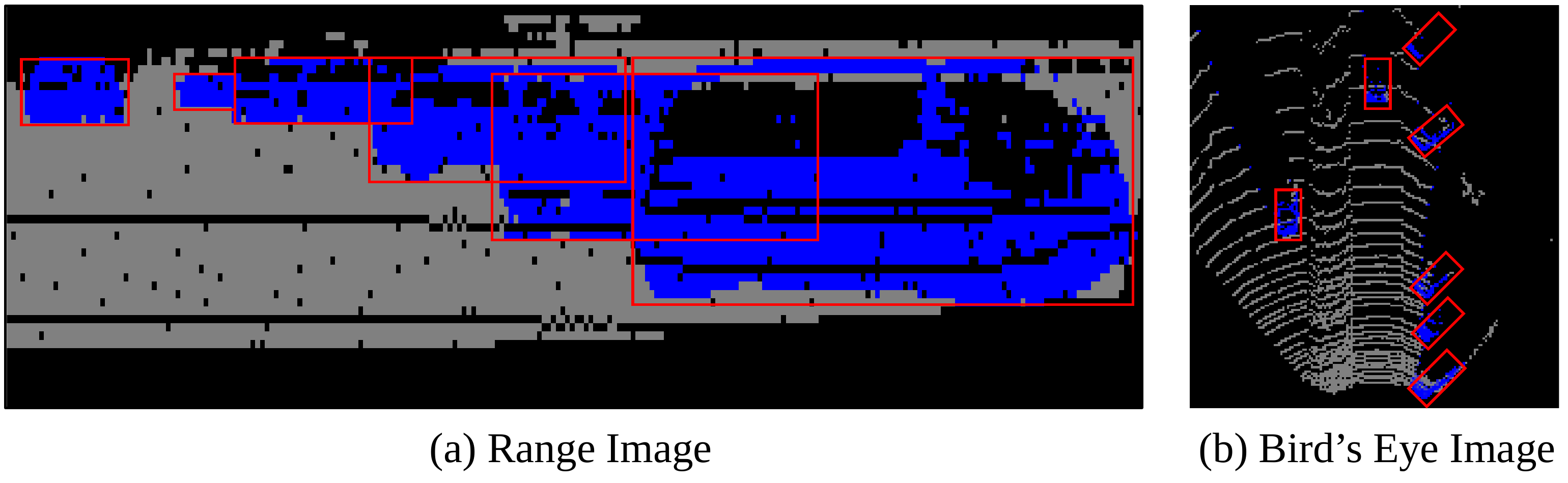}
		\caption{Comparison of the range image~(left) and the bird's eye image~(right). The bounding boxes in the range image differs notably in scale and easily overlap with each other. In contrast, the bounding boxes in the bird's eye image maintain similar size and do not have overlapping areas. }
		\label{rvbev}
	\end{figure}
	
	\subsection{RV-PV-BEV Module} \label{sec_rpe}
	The range image representation is suitable for feature extraction by utilizing the 2D convolution. However, it is difficult to assign anchors in the range image plane due to the large scale variation. The severe occlusion also makes it difficult to remove redundant bounding boxes in the Non-Maximum Suppression~(NMS) module. Fig.~\ref{rvbev} shows a typical example. The size of the bounding boxes varies greatly in the range image. Some bounding boxes have a large overlap. In contrast, these bounding boxes have a similar shape in the BEV plane because most cars are of similar size. It is also impossible for different cars to overlap with each other in the BEV place even though they are very close. We therefore think that it is more suitable to generate anchors in the BEV plane. Thus, we transfer the feature extracted from the range image to the bird's eye image.
	
	For each point, we record its corresponding pixel coordinates in the range image plane, so we can obtain the pointwise feature by indexing the output feature of the range image backbone. Then, we project the pointwise feature to the BEV plane. For points corresponding with the same pixel in the BEV image, we use the average pooling operation to generate the representative feature for the pixel. Here the point view only serves as the bridge to transfer features from the range image to the BEV image. We do not use the point-based convolution to extract features from points.
	
	\textbf{Discussion. }Different from projecting point clouds to the BEV plane at the beginning of the network~\cite{chen2017mv3d, ku2018avod}, we perform the projection after extracting high-level features. If projecting at the beginning, the BEV serves as the main feature extractor. The information loss caused by the discretization leads to inaccurate features. In our framework, the BEV primarily plays the role of anchor generation. As we have extracted features from the lossless range image, the quantization error caused by the discretization has a minor influence. Experiments show the superiority of our methods compared with those methods projecting at the beginning.

	\subsection{3D RoI Pooling} \label{sec_pooling}
\begin{figure}[t]
		\centering
		\includegraphics[width=1.0\linewidth]{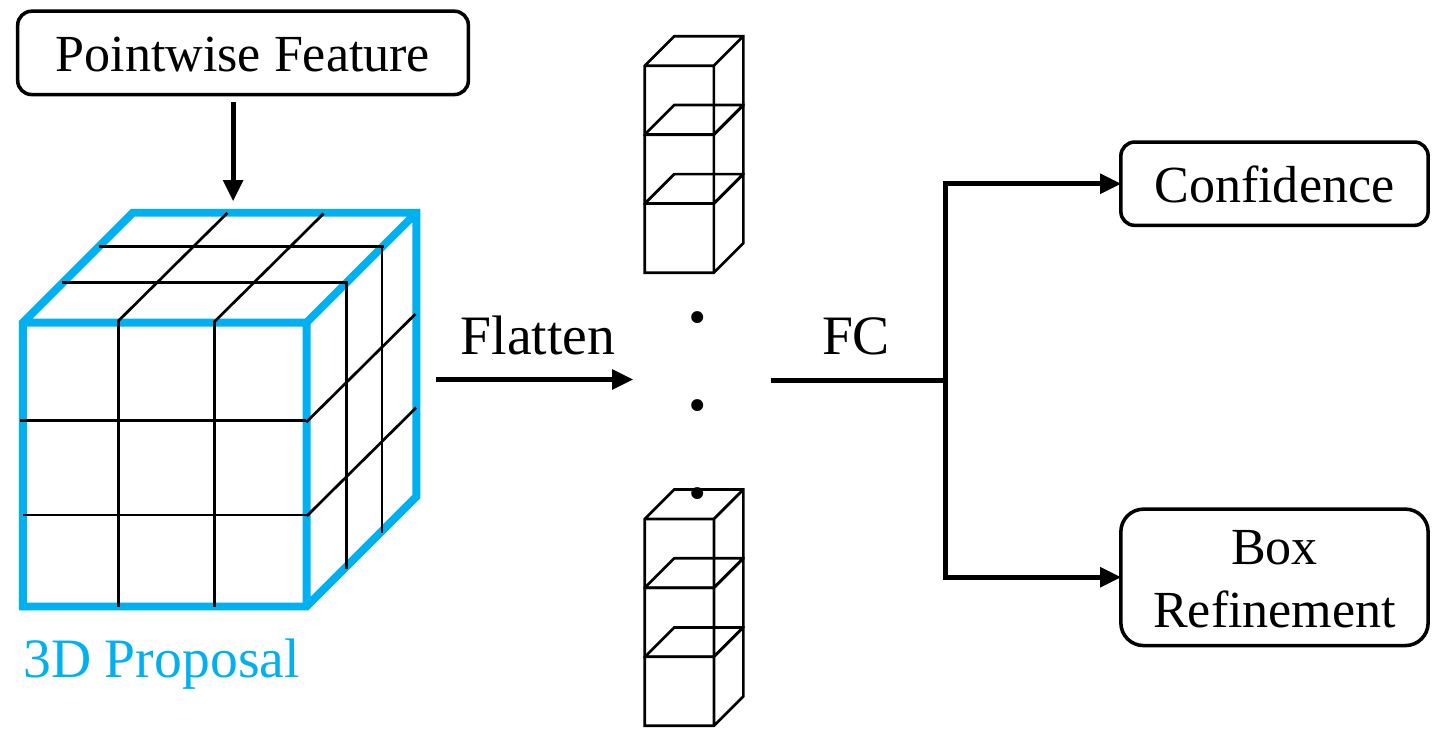}
		\caption{Illustration of the 3D RoI pooling. The blue box is the 3D proposal. It is divided into the regular grids along three aligned axes. Each grid obtains the feature from the aforementioned point view. The max pooling operation is applied to pool multiple point features within a grid. All 3D grids are flattened to a vectorized feature. Several fully connected layers are applied to predict the refined boxes and the confidences.}
		\label{roipool}
	\end{figure}

	Based on the bird's eye image, we generate 3D proposals using the region proposal network~(RPN). However, neither the range image nor the bird's eye image explicitly learns features along the height direction of the 3D bounding box, which causes our predictions to be relatively accurate in the BEV plane, but not in the 3D space. As a result, we want to explicitly utilize the 3D space information. We conduct a 3D RoI pooling based on the 3D proposals generated by RPN. The proposal is divided into a fixed number of grids. Different grids contain different parts of the object. As these grids have a clear spatial relationship, the height information is encoded among their relative positions. We directly vectorize these grids from three dimensions to one dimension sorted by their 3D positions~(illustrated in Fig.~\ref{roipool}). We apply several fully connected layers to the vectorized features and predict the refined bounding boxes and the corresponding confidences.
	
	\subsection{Loss Function} \label{sec_loss}
	The whole network is trained in end-to-end fashion. The loss contains two main parts: the region proposal network loss $L_{rpn}$ and the region convolutional neural network loss $L_{rcnn}$, which is similar to~\cite{yan2018second,shi2019part,shi2020pv}. The RPN loss $L_{rpn}$ includes the focal loss $L_{cls}$ for anchor classification, the smooth-L1 loss $L_{reg}$ for anchor regression and the direction classification loss $L_{dir}$ for orientation estimation~\cite{yan2018second}:
	\begin{equation}
	L_{rpn} = L_{cls} + \alpha L_{reg} + \beta L_{dir}
	\label{rpn_loss}
	\end{equation}
	where $\alpha$ is set to 2 and $\beta$ is set to 0.2 empirically. We use the default parameters for focal loss~\cite{lin2018focal}. The smooth-L1 loss regresses the residual value relative to the anchor~\cite{yan2018second}.
	
	The RCNN loss includes the confidence prediction loss $L_{score}$ guided by the IoU~\cite{shi2019part}, the smooth-L1 loss $L_{reg}$ for refining proposals and the corner loss $L_{corner}$~\cite{qi2018fpoint}:
	\begin{equation}
	L_{rcnn} = L_{score} + L_{reg} + L_{corner}
	\label{rcnn_loss}
	\end{equation}
	
	The total training loss is the sum of the above losses:
	\begin{equation}
	L_{total} = L_{rpn} + L_{rcnn}
	\label{total_loss}
	\end{equation}

	\begin{figure*}[!tb]
		\centering
		\includegraphics[width=0.99\linewidth]{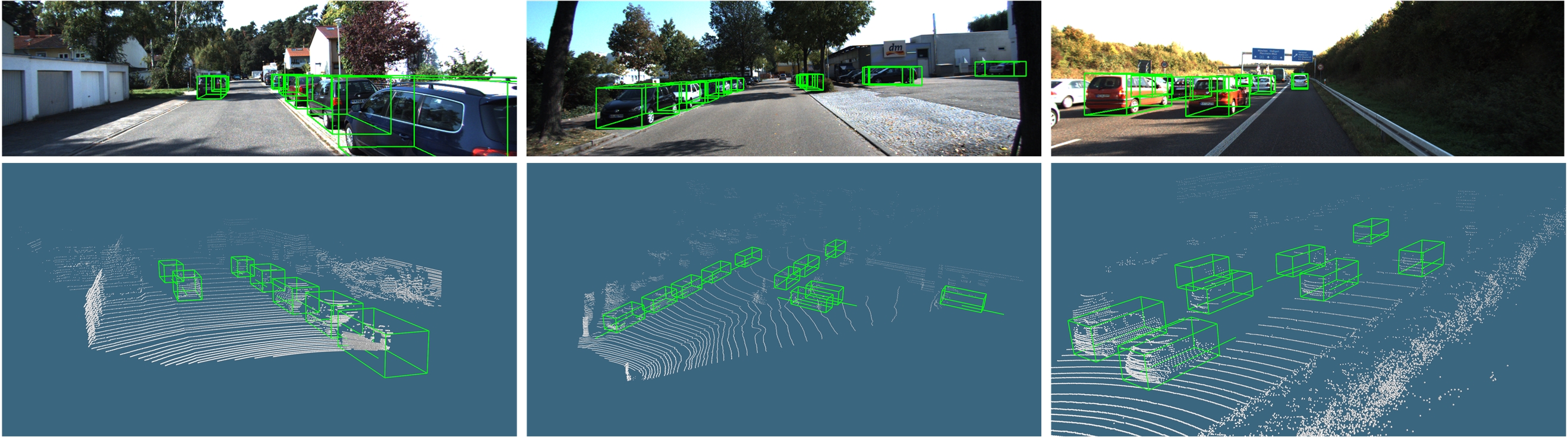}
		\caption{Visualization of our predictions on the KITTI dataset.}
		\label{result}
	\end{figure*}
	 
	\newcommand{\tabincell}[2]{\begin{tabular}{@{}#1@{}}#2\end{tabular}}  
	\begin{table*}[tb]
		\normalsize
		\caption{Performance comparison with previous methods on the KITTI online test server. The AP with 40 recall positions~($R40$) is used to evaluate the 3D object detection and BEV object detection. }
		\label{KITTI_test}
		\begin{center}
			\resizebox{1.0\textwidth}{!}{
				\begin{tabular}{ccccccccc}
					\toprule
					\multirow{2}*{Method} &\multirow{2}*{Reference} & \multirow{2}*{Modality} & \multicolumn{3}{c}{3D} & \multicolumn{3}{c}{BEV}\\
					~ & ~ & ~ & Easy & Moderate & Hard & Easy & Moderate & Hard \\
					\midrule
					\textbf{Multi-Modality: }\\
					MV3D~\cite{chen2017mv3d} 			& CVPR 2017 & RGB+LIDAR & 74.97 & 63.63 & 54.00 & 86.62 & 78.93 & 69.80 \\
					ContFuse~\cite{liang2018deep} 		& ECCV 2018 & RGB+LIDAR & 83.68 & 68.78 & 61.67 & 94.07 & 85.35 & 75.88 \\
					AVOD-FPN~\cite{ku2018avod} 		& IROS 2017 & RGB+LIDAR & 83.07 & 71.76 & 65.73 & 90.99 & 84.82 & 79.62 \\
					F-PointNet~\cite{qi2018fpoint} 		& CVPR 2018 & RGB+LIDAR & 82.19 & 69.79 & 60.59 & 91.17 & 84.67 & 74.77 \\
					UberATG-MMF~\cite{Liang_2019_CVPR} 	& CVPR 2019 & RGB+LIDAR & 88.40 & 77.43 & 70.22 & 93.67 & 88.21 & 81.99 \\
					EPNet~\cite{huang2020epnet}			& ECCV 2020 & RGB+LIDAR & 89.91 & 79.28 & 74.59 & 94.22 & 88.47 & 83.69 \\
					\midrule
					\textbf{LIDAR-only: }\\
					PointRCNN~\cite{shi2019pointrcnn} 		& CVPR 2019 & Point & 86.96 & 75.64 & 70.70 & 92.13 & 87.39 & 82.72 \\
					Point-GNN~\cite{shi2020point}		& CVPR 2020 & Point & 88.33 & 79.47 & 72.29 & 93.11 & 89.17 & 83.90 \\
					3D-SSD~\cite{yang20203dssd}			& CVPR 2020 & Point & 88.36 & 79.57 & 74.55 & 92.66 & 89.02 & 85.86 \\
					SECOND~\cite{yan2018second} 			& Sensors 2018 & Voxel & 83.34 & 72.55 & 65.82 & 89.39 & 83.77 & 78.59 \\
					PointPillars~\cite{Lang_2019_CVPR} 	& CVPR 2019 & Voxel & 82.58 & 74.31 & 68.99 & 90.07 & 86.56 & 82.81 \\
					3D IoU Loss~\cite{zhou2019iou} 	& 3DV 2019 & Voxel & 86.16 & 76.50 & 71.39 & 91.36 & 86.22 & 81.20 \\
					Part-A2~\cite{shi2019part} 		& TPAMI 2020 & Voxel & 87.81 & 78.49 & 73.51 & 91.70 & 87.79 & 84.61 \\
					Fast Point R-CNN~\cite{Chen_2019_ICCV} & ICCV 2019 & Voxel+Point & 85.29 & 77.40 & 70.24 & 90.87 & 87.84 & 80.52 \\
					STD~\cite{yang2019std} 			& ICCV 2019 & Voxel+Point & 87.95 & 79.71 & 75.09 & 94.74 & 89.19 & \textbf{86.42} \\
					SA-SSD~\cite{he2020structure}			& CVPR 2020  & Voxel+Point & 88.75 & 79.79 & 74.16 & \textbf{95.03} & \textbf{91.03} & 85.96 \\
					PV-RCNN~\cite{shi2020pv} 		& CVPR 2020 & Voxel+Point & \textbf{90.25} & \textbf{81.43} & 76.82 & 94.98 & 90.65 & 86.14 \\
					LaserNet~\cite{Meyer_2019_CVPR} 		& CVPR 2019 & Range & - & - & - & 79.19 & 74.52 & 68.45 \\
					\midrule
					RangeRCNN~(ours) & - & Range & 88.47 & 81.33 & \textbf{77.09} & 92.15 & 88.40 & 85.74\\
					
					\bottomrule
				\end{tabular}
				
			}
		\end{center}
	\end{table*}
	
	\begin{table}[!t]
		\normalsize
		\caption{Runtime analysis of state-of-the-art two-stage detectors.}
		\label{runtime_analysis_2}
		\begin{center}
			\resizebox{0.45\textwidth}{!}{
				
				\begin{tabular}{ccccc}
					\toprule
					~ & PointRCNN & PartA2 & PV-RCNN & RangeRCNN \\
					\midrule
					FPS~(Hz) & 12 & 13 & 10 & 22 \\
					\bottomrule
				\end{tabular}
				
			}
		\end{center}
	\end{table}
	
	\section{EXPERIMENTS}
	In this section, we evaluate our proposed RangeRCNN on the challenging KITTI dataset~\cite{geiger2012we} and the large-scale Waymo Open dataset~\cite{sun2020scalability}.
	
	\textbf{Dataset. }
	The KITTI dataset~\cite{geiger2012we} contains 7481 training samples and 7518 test samples. We follow the general split of 3712 training samples and 3769 validation samples. The KITTI dataset provides a benchmark for 3D object detection. We compare our proposed RangeRCNN with other state-of-the-art methods on this benchmark.
	
	The Waymo Open dataset~\cite{sun2020scalability} provides 798 training sequences and 202 validation sequences. Each sequence contains 200 frames within 20 seconds. Specifically, this dataset provides the point clouds in the form of the range image, so we directly obtain the raw range image and do not need to convert the point clouds to the range image.

	\subsection{Implementation Details}
	\textbf{Network Details. }
	For the KITTI dataset, the size of the input range image is $5\times 48\times 512$ as mentioned above. For the Waymo Open dataset, the size of the range image is $6\times 64\times 2650$, where 64 indicates 64 lasers and 2650 means that each laser has 2650 points. The LIDAR of the Waymo Open dataset has a 360 degree field of view instead of 90 degrees as the KITTI dataset. The range image backbone for the two datasets is described in Sec.~\ref{sec_backbone}.
	
	For the KITTI dataset, the range of BEV is within $[0, 69.12]m$ for the x-axis and $[-39.68, 39.68]m$ for the y-axis. The resolution of BEV is $0.16^{2}~m^{2}$, so the initial spatial size of BEV is $496\times 432$. We use three convolution blocks to downsample the BEV to $248\times 216$, $124\times 108$ and $62\times 54$, and upsample the three sizes to $248\times 216$. The three features with the same size are then concatenated along the channel axis. For the Waymo Open dataset, the range of BEV is within $[-75.52, 75.52]m$ for the x-axis and $[-75.52, 75.52]m$ for the y-axis. The resolution of BEV is $0.32^{2}~m^{2}$, so the initial spatial size of BEV is $472\times 472$, which is slightly different from the size of $496\times 432$ for the KITTI dataset. The aforementioned concatenated features are input to the RPN network for generating 3D proposals.
		
	In the two-stage RCNN, the 3D proposal generated by the RPN is divided into a fixed number of grids along with the local coordinate system of the 3D proposal. The spatial shape is set as $12\times12\times12$. We reshape the $12\times12\times12$ grids to a vectorized format with dimension of $12^{3}\times C$, where $C$ is the feature dimension of each grid. The feature of each grid is obtained from the point features. $C$ is equal to 64. If multiple points fall within the same grid, the max pooling is used. Three fully connected layers are then applied to the vectorized feature. Finally, the confidence branch and the refinement branch are used to output the final result.

	\begin{table*}[!t]
		\normalsize
		\caption{Performance comparison with previous methods on the level 1 vehicle of the Waymo validation set.}
		\label{waymo_dataset}
		\begin{center}
			\resizebox{0.8\textwidth}{!}{
				\begin{tabular}{ccccccccc}
					\toprule
					~ & \multicolumn{4}{c}{3D mAP} & \multicolumn{4}{c}{3D mAPH} \\
					\cmidrule(r){2-5} \cmidrule(r){6-9}
					~ & All & 0-30m & 30-50m & 50-75m & All & 0-30m & 30-50m & 50-75m \\
					\midrule
					LaserNet~\cite{Meyer_2019_CVPR} & 52.11 & 70.94 & 52.91 & 29.62 & 50.05 & 68.73 & 51.37 & 28.55\\
					PointPillars~\cite{Lang_2019_CVPR} & 56.62 & 81.01 & 51.75 & 27.94 & - & - & - & -\\
					RCD~\cite{bewley2020range} & 68.95 & 87.22 & 66.53 & 44.53 & 68.52 & 86.82 & 66.07 & 43.97\\
					PV-RCNN~\cite{shi2020pv} & 70.30 & 91.92 & 69.21 & 42.17 & 69.69 & 91.32 & 68.53 & 41.31\\
					RangeRCNN~(ours) & \textbf{75.43} & \textbf{92.01} & \textbf{74.27} & \textbf{54.23} & \textbf{74.97} & \textbf{91.52} & \textbf{73.77} & \textbf{53.54}\\
					\bottomrule
				\end{tabular}
				
			}
		\end{center}
		\arrayrulecolor{black}
	\end{table*}

	\textbf{Training and Inference Details. }
	We implement the proposed RangeRCNN with PyTorch1.3. The network can be trained in an end-to-end fashion with the ADAM optimizer. For the KITTI dataset, we train the entire network with the batch size of 32 and learning rate of 0.01 for 80 epochs on 8 NVIDIA Tesla V100 GPUs, which takes approximately 1.5 hours. For the large-scale Waymo Open dataset, we train the entire network with the batch size of 32 and learning rate of 0.01 for 30 epochs on 8 NVIDIA Tesla V100 GPUs, which takes approximately 70 hours. We adopt the cosine annealing learning rate strategy for the learning rate decay.
	
	For the KITTI dataset, we use the data augmentation strategy similar to~\cite{yan2018second,shi2019part}, including random flipping along the x-axis, random global scaling with a scaling factor sampled from $[0.95, 1.05]$, random global rotation around the vertical axis with a sampled angle from $[-\frac{\pi}{4},\frac{\pi}{4}]$, and ground-truth sampling augmentation to randomly "paste" some new ground-truth objects to current training scenes. For the Waymo Open dataset, we discuss the data augmentation strategy in Sec.~\ref{discuss_aug}.
	
	For RCNN, we choose 128 proposals with a 1:1 ratio for positive and negative samples during training. During inference, we retain the top 100 proposals according to the confidence with the NMS threshold 0.7. We apply the 3D NMS to the refined bounding boxes with a threshold of 0.1 to generate the final result.

	\subsection{3D Detection on the KITTI Dataset}
	\textbf{Metrics. }
	The detection result is evaluated using the mean average precision~(mAP) with the IoU threshold 0.7. For the official test benchmark, the mAP with 40 recall positions is used. We report the performance of 3D detection and BEV detection in Table~\ref{KITTI_test}.
	
	\textbf{Performance. }
	We submit our results to the online benchmark for comparison with other state-of-the-art methods. For evaluating the test set, we use all provided labeled samples to train our model. Table~\ref{KITTI_test} reports the results evaluated on the KITTI online test server. Fig.~\ref{result} shows the qualitative results. Our RangeRCNN outperforms almost all previous approaches except PV-RCNN~\cite{shi2020pv} on the commonly used moderate level for 3D car detection. We surprisingly observe that our method achieves the highest accuracy for the hard level. We think that the performance is beneficial to two aspects. First, some hard examples are very sparse in the 3D space, but they have more obvious features in the range image thanks to the compact representation. These objects can thus be detected using the range image representation. Second, RCNN further refines the 3D position of the bounding box, which boosts the 3D performance. The ablation study also proves the value of RCNN.

	\textbf{Runtime Analysis.}
	To provide a fair runtime comparison, we run several representative two-stage methods with the same device~(NVIDIA V100 GPU). The comparison is shown in Table.~\ref{runtime_analysis_2}. The proposed RangeRCNN runs at 22 fps, which is much faster than the point-based PointRCNN and the voxel-based PartA2 and PV-RCNN. The high computation performance is beneficial from the compact representation of the range image and the efficiency of 2D convolution. Research on the range image can promote real-time 3D object detection.

	\subsection{3D Detection on the Waymo Open Dataset} \label{sec_waymo}
	\textbf{Metrics. }
	The mean average precision~(mAP) and the mean average precision weighted by heading~(mAPH) are adopted as the metrics of this dataset. The IoU threshold for vehicles is set as 0.7. The detection performances with different object distances are reported. The dataset divides the objects into two difficulty levels based on the number of points within the 3D boxes. Level 1 includes at least five points with the 3D box and Level 2 includes at least one point. As few previous methods report their Level 2 performance, we compare RangeRCNN with state-of-the-art methods with respect to Level 1.
	
	\textbf{Performance. }
	Table~\ref{waymo_dataset} shows that RangeRCNN outperforms previous methods by a large margin. The Waymo dataset directly provides the range image without the requirement of additional conversion, so the quality of the range image is better than that of the KITTI dataset. RangeRCNN has the ability to extract rich information for distant objects. For objects within $[0, 30]m$, the performance of RangeRCNN is similar to that of PV-RCNN. For objects within $[30, 50]m$ and $[50, 75]m$, RangeRCNN performs much better than PV-RCNN. The farther away the object is, the greater the advantage of our method over other methods. RangeRCNN outperforms RCD which also utilizes the range image representation. RCD only operates on the range image which does not utilize the ability of BEV to generate anchors. The performance of RangeRCNN on this large-scale dataset shows the large potential of the range image based method.
	
	\begin{figure*}[t]
		\centering
		\includegraphics[width=1.0\linewidth]{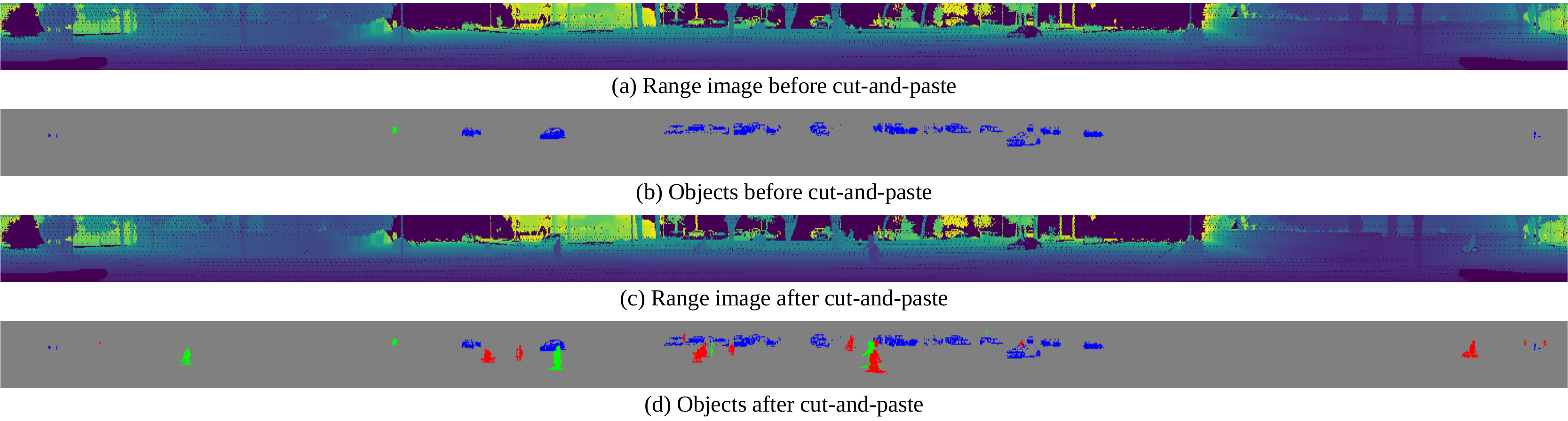}
		\caption{Illustration of the cut-and-paste strategy for the Waymo Open dataset. Vehicles, Pedestrians and Cyclists are marked blue, green and red, respectively. }
		\label{cutandpaste}
	\end{figure*}

	\subsection{Discussion on Data Augmentation of Range Image.} \label{discuss_aug}
	For the KITTI dataset, it does not directly provide the range image, so we need to convert the points to the range image. In this dataset, we first perform data augmentation on the points similar with~\cite{yan2018second, shi2019part}. Then we convert the augmented points to the range image. Although the augmented range image no longer strictly adheres to the scanning characteristic of the rotating LIDAR, the augmentation strategy brings an improvement in accuracy.
	
	For the Waymo Open dataset, the data augmentation pipeline is a little bit different from that of the KITTI dataset. Since the Waymo dataset provides the range image representation, we directly flip and rotate the range image, and scale the point coordinates stored in the channel of the range image. One interesting operation is to apply the cut-and-paste strategy on the range image~(illustrated in Fig.~\ref{cutandpaste}).
	We first traverse the training data and read all annotated objects. For each point in the object, we store its 3D coordinate, intensity, elongation and corresponding pixel coordinate on the range image plane. When training the model, we randomly paste the object according to the pixel coordinates to augment the original range image. In addition to augment the range image, we also need to simultaneously consider the corresponding changes and transformation for the 3D point coordinates. The improvement brought by the cut-and-paste strategy depends on several factors. If we use a small part of the whole Waymo Open dataset for training, the improvement will be more obvious, which means that this strategy is very effective with limited training data. For the category with much fewer labels such as cyclists, the improvement will be much greater, which means that this strategy is helpful for some scarce categories. As the Waymo Open dataset is large, if training with all data, this strategy hardly improves the accuracy of vehicles with large amount of annotations. We believe that the augmentation strategy is very useful for future reserach based on range image. Through the range image based data augmentation, the preprocessing pipeline of the range image based method is almost equivalent to the voxel based method and the point based method. Detailed experiments and analysis will be provided in the future version.

	\begin{table}[!t]
		\normalsize
		\caption{Comparison of the one-stage model RangeDet and the two-stage model RangeRCNN. The AP with 40 recall positions~(R40) is used.}
		\label{ablation_roi}
		\begin{center}
			\resizebox{0.5\textwidth}{!}{
				
				\begin{tabular}{ccccccc}
					\toprule
					\multirow{2}*{Method} & \multicolumn{3}{c}{3D} & \multicolumn{3}{c}{BEV} \\
					~ & Easy & Moderate & Hard & Easy & Moderate & Hard \\
					\midrule
					RangeDet & 89.87 & 80.72 & 77.37 & 92.07 & 88.37 & 87.03 \\
					RangeRCNN & \textbf{91.41} & \textbf{82.77} & \textbf{80.39} & \textbf{92.84} & \textbf{88.69} & \textbf{88.20} \\
					\bottomrule
				\end{tabular}
				
			}
		\end{center}
	\end{table}
		
	\begin{table}[!t]
		\normalsize
		\caption{Comparison of different pooling sizes in \{6, 8, 10, 12, 14\}. The AP with 40 recall positions~(R40) is used.}
		\label{ablation_roisize}
		\begin{center}
			\resizebox{0.5\textwidth}{!}{
				
				\begin{tabular}{ccccccc}
					\toprule
					\multirow{2}*{Method} & \multicolumn{3}{c}{3D} & \multicolumn{3}{c}{BEV} \\
					~ & Easy & Moderate & Hard & Easy & Moderate & Hard \\
					\midrule
					RangeRCNN-6 & 89.55 & 82.33 & 80.01 & 92.56 & 88.47 & 87.78 \\
					RangeRCNN-8 & 89.23 & 82.35 & 79.96 & 92.44 & 88.49 & 88.00 \\
					RangeRCNN-10 & 89.48 & 82.62 & 80.36 & 92.76 & 88.60 & 88.11 \\
					RangeRCNN-12 & 91.41 & \textbf{82.77} & \textbf{80.39} & \textbf{92.84} & \textbf{88.69} & \textbf{88.20} \\
					RangeRCNN-14 & \textbf{91.54} & 82.61 & 80.29 & 92.67 & 88.49 & 88.16 \\
					\bottomrule
				\end{tabular}
				
			}
		\end{center}
	\end{table}
	\subsection{Ablation Study}
	In this section, we conduct the ablation study on the validation set of the KITTI dataset.
	
	\textbf{Effects of 3D RoI Pooling. }
	As the entire framework is two-stage, we compare the results of the one-stage model RangeDet and the two-stage model RangeRCNN to analyze the value of RCNN. From Table~\ref{ablation_roi}, we can find that RangeDet and RangeRCNN exhibit similar performances for BEV detection. For 3D detection, RangeRCNN outperforms RangeDet by a large margin. The better 3D performance comes from the 3D information encoded by 3D RoI pooling. 
	
	\textbf{Effects of Grid Size in 3D RoI Pooling. }
	We further evaluate the influence of the grid size in 3D RoI pooling. We compare a set of grid sizes in \{6, 8, 10, 12, 14\}. Table~\ref{ablation_roisize} shows the results. It can be found that the grid size exerts no great influence on the metric. We choose 12 as the grid size, which is a relatively better size.

	\section{CONCLUSIONS}
	In this paper, we explore the potential of the range image representation and present a novel framework called RangeRCNN for fast and accurate 3D object detection. By combining the advantages of extracting lossless features on the range image and generating high-quality anchors on the bird's eye image, our method achieves state-of-the-art performance on the KITTI dataset and the Waymo Open dataset while maintaining high efficiency. The compact representation of the range image provides more possibilities for real-time 3D object detection in large outdoor scenes.
	
	
	
	\bibliographystyle{IEEEtran}
	\bibliography{IEEEabrv,bibtex/reference}

\end{document}